\pdfoutput=1

\documentclass[11pt]{article}
\usepackage[preprint]{acl}
\usepackage{times}
\usepackage{latexsym}
\usepackage{amsfonts}
\usepackage[T1]{fontenc}
\usepackage[utf8]{inputenc}
\usepackage{microtype}
\usepackage{inconsolata}
\usepackage{graphicx}

\title{Memory Tokens: Large Language Models Can Generate Reversible Sentence Embeddings}

\author{Ignacio Sastre \and Aiala Rosá\\\\
Instituto de Computación, Facultad de Ingeniería, Universidad de la República \\ Montevideo, Uruguay\\\texttt{\{isastre,aialar\}@fing.edu.uy}}

\begin{document}
\maketitle
\begin{abstract}
In this work, we observe an interesting phenomenon: it is possible to generate reversible sentence embeddings that allow an LLM to reconstruct the original text exactly, without modifying the model’s weights.
This is achieved by introducing a special memory token, whose embedding is optimized through training on a fixed sequence.
When prompted with this embedding, the model reconstructs the fixed sequence exactly.
We evaluate this phenomenon across English and Spanish datasets, sequences of up to approximately 240 tokens, and model scales ranging from 100M to 8B parameters.
Notably, Llama 3.1 8B successfully reconstructs all tested sequences. 
Our findings highlight an interesting capability of LLMs and suggest potential applications in memory-based retrieval, compression, and controlled text generation.\footnote{Code repo with the implementation: \url{https://github.com/nsuruguay05/memory_token}}
\end{abstract}

\section{Introduction}

Large Language Models (LLMs) encode textual information in high-dimensional embeddings, capturing semantic and syntactic structures.

In this work, we observe and explore an interesting phenomenon using LLMs: it is possible to construct reversible embeddings that encode an arbitrary text sequence in such a way that the original text can be perfectly reconstructed when used as input to an LLM, without modifying the model's weights.

This reversibility emerges when training a dedicated embedding associated with a special token, which we call a \textit{memory token}, on a fixed sequence.
By overfitting this embedding to a given text while keeping the model frozen, we show that the same model can autoregressively reconstruct the text when prompted with the learned embedding.

We study this phenomenon, evaluating its effectiveness across different domains, sequence lengths, and languages (English and Spanish).
Additionally, we explore models of various sizes, including GPT-2~\cite{radford2019language} and the Llama 3 family~\cite{llama3modelcard}.

This observation sheds light on the representational capacity of LLMs and opens up new possibilities for memory-based retrieval and controlled text generation.
It also suggests potential applications in text compression, adversarial attacks, and interpretability research.

\begin{figure*}[t]
  \centering
    \includegraphics[width=\linewidth]{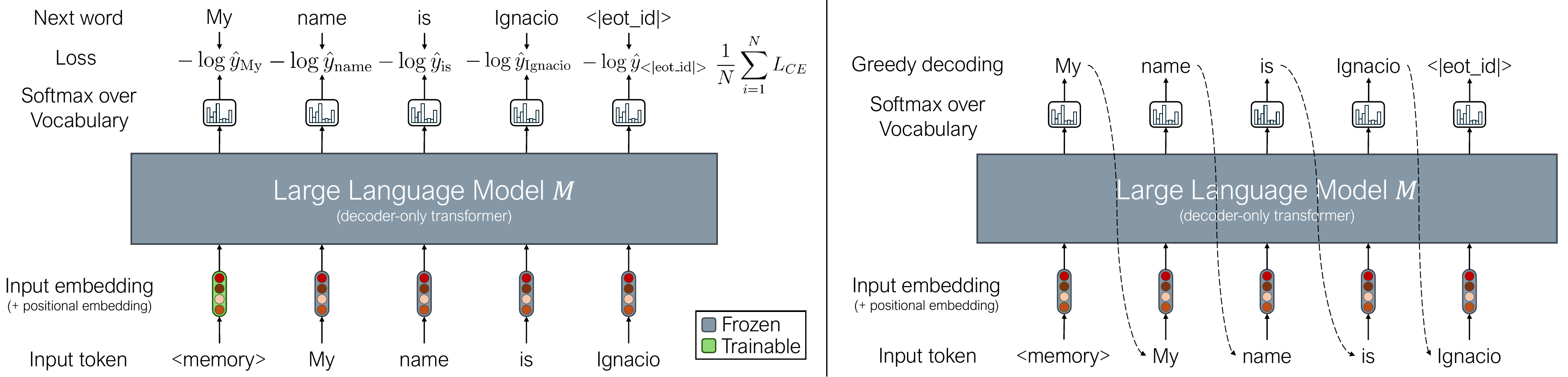}
  \caption{Left: Illustration of the training process. Only the embedding corresponding to the memory token is trainable. Right: Illustration of inference with the memory token as input. Following a greedy decoding strategy, the model reconstructs the original text.}
  \label{fig:train_inference}
\end{figure*}

\section{Related work}

The method of optimizing a set of vectors and using them as a prefix for a specific task belongs to a class of techniques known as P*-tuning~\cite{li-liang-2021-prefix}.
Some of these techniques are mentioned below.

Prefix tuning~\cite{li-liang-2021-prefix} applies this idea as a lightweight alternative to full fine-tuning.
It involves prepending a sequence of task-specific vectors to the input, optimizing these vectors while keeping the model frozen.

Building on Prefix tuning, Prompt-tuning~\cite{lester-etal-2021-power} applies the same principles and demonstrates the importance of scale: larger models get better results with this method and even become competitive with full fine-tuning.

In a similar way, P-Tuning~\cite{liu-2024-ptuning} is proposed as a method to improve the performance of discrete prompting.
It employs trainable continuous prompt embeddings in concatenation with discrete prompts.

More broadly, P*-tuning methods can be seen as a type of soft prompting in the context of prompt compression~\cite{li-etal-2025-prompt}.
Soft prompt methods aim to compress text into a smaller number of special tokens.
In this context, similar to our approach, \citet{wingate-etal-2022-prompt} propose training these embeddings using contrastive conditioning, without modifying the model's weights.
While they focus on prompt compression, we aim to demonstrate the ability of LLMs to exactly reconstruct the original sequence.
Moreover, our work differs in how we train the embedding: we essentially overfit it to a single sequence.

A more recent work in this line is 500xCompressor~\cite{li2024500xcompressorgeneralizedpromptcompression}.
It uses an encoder-decoder setup with LoRA parameters in the encoder and trains to encode textual information into the key-value (KV) pairs of compressed tokens.
500xCompressor is designed for prompt compression to improve LLM efficiency in downstream tasks such as question answering, using a two-step training pipeline (pretraining and fine-tuning).
In contrast, our work uses a decoder-only LLM and directly optimizes a single embedding vector, without relying on an encoder.

There has also been work on creating reversible sentence embeddings.
\citet{kugler-2024-invbert} propose a method for reconstructing text from contextualized embeddings generated by BERT~\cite{devlin-etal-2019-bert}. 
Their approach involves training a decoder model to recover the original text given the contextualized embedding as input.

\citet{li-etal-2023-sentence} follow a similar approach to the one proposed in this work.
They use already existing sentence embedding models to generate an embedding, which is then used as the initial input vector for a decoder model.
The decoder is fine-tuned to reconstruct the original sequence.
However, our approach differs in that we do not fine-tune the LLM or rely on pre-trained sentence embedding models.

\section{Memory token}

We define a \textit{memory token} as a new token added to the model's vocabulary.
This token has a corresponding embedding in the LLM's embedding layer, that serves as a dense vector representation of an arbitrary text sequence.

These embeddings are reversible, meaning that the original text sequence can be reconstructed from them, allowing the memory token to effectively store and retrieve the sequence it encodes.

\subsection{Training}

A new token, \texttt{<MEMORY>}, is added to the model's vocabulary with a randomly initialized embedding in the LLM’s embedding layer.
A sequence of tokens $x = x_1, x_2, ...,x_N$ is defined to be used for training, following the template: \texttt{“<MEMORY>\{text\}<|eot\_id|>”}, where \texttt{text} represents the text to be memorized and \texttt{<|eot\_id|>} is the EOS token of the model.

The entire model is frozen, including its embedding layer, with the only exception of the \texttt{<MEMORY>} token's embedding.
This is the only set of parameters that is updated during training.

The training objective is the standard cross-entropy loss used for autoregressive generation. Given the input sequence $x$, the model is trained to predict each token $x_t$ conditioned on the preceding tokens $x_{<t}$. Formally, the loss function is defined as:
$$\mathcal{L} = -\frac{1}{N} \sum_{t=1}^{N} \log P(x_t \mid x_{<t}; \theta)$$
where $N$ is the sequence length, and $P(x_t \mid x_{<t}; \theta)$ represents the model’s predicted probability of token $x_t$.
Figure \ref{fig:train_inference} (left) demonstrates an example of this process.

Training is performed by repeatedly optimizing the embedding using the same sequence $x$ until the model generates the expected output exactly or a maximum number of iterations is reached.
This process effectively overfits the embedding to the given sequence, ensuring that it precisely encodes the target text sequence.

This process must be repeated using a new memory token for each new sequence that needs to be learned.

\subsection{Inference}

The resulting embedding can be extracted and used as a dense representation of the input text.
More interestingly, if training is stopped before reaching the maximum iterations,  when this embedding is provided as input to the same model it was trained with, and a greedy decoding strategy is applied (selecting the highest probability token at each step), the model generates the original sequence word by word.
Figure \ref{fig:train_inference} (right) illustrates this process.

It is important to note that the model itself remains unchanged after the training phase.
This implies that the learned embedding encodes a representation that forces the model to generate the exact desired text.

\section{Experiments}

To demonstrate this phenomenon and evaluate its scope and generalization, we construct datasets with varying characteristics, including different domains, sequence lengths, and languages.
We then measure the ability of different models to reconstruct these sequences effectively.

We evaluate the effectiveness of reconstruction using accuracy, which we define as the proportion of correctly predicted tokens with respect to the original sequence $x$. 
For each predicted token $\hat{x}_t$, the correct prefix $x_{<t}$ is given to the model.
The accuracy for a given sequence is then computed as:
$$ Acc(\hat{x}, x) = \frac{1}{N}\sum_{t=1}^{N}\mathbb{I}\{\hat{x}_t = x_t\}$$
where $\mathbb{I}\{\hat{x}_t = x_t\}$ is an indicator function that equals 1 if the predicted token $\hat{x}_t$ matches the ground truth token $x_t$, and 0 otherwise.

All experiments were conducted with a maximum of 3000 iterations.
For the smaller models, a linear learning rate scheduler was employed, initializing the learning rate at 0.2 and increasing it linearly to 1.0 by the 100th iteration.
For Llama 3.1 8B, a higher learning rate yielded better results; thus, we report experiments using a learning rate of 5.0.

Table~\ref{tab:models} provides an overview of the LLMs used in these experiments.
We selected models of varying sizes to determine the impact of scale on this phenomenon.
As shown in the table, smaller models typically have lower-dimensional embeddings, which is an important factor to consider since the entire sequence must be encoded within a single embedding.

\subsection{Datasets}

\paragraph{Sebastian Raschka blog}

To ensure that the text is not present in the training corpus of the models, we selected a recent blog post from Raschka’s blog, \textit{New LLM Pre-training and Post-training Paradigms}\footnote{\url{https://sebastianraschka.com/blog/2024/new-llm-pre-training-and-post-training.html}}.
We segmented the blog post into non-overlapping chunks of 100 and 1000 characters, creating two datasets with varying sequence lengths.
For efficiency, we used only the first 20 sequences from each dataset.

\paragraph{Faculty of Engineering chunks}

This corpus consists of Spanish text chunks extracted from the Faculty of Engineering website at the Universidad de la República.
These chunks were originally used in a Retrieval-Augmented Generation (RAG) system and present various challenges, such as embedded YouTube links and named entities.
They were manually created and have an average length of 681 characters. As with the previous datasets, we only used the first 20 chunks for this evaluation.\\

Table~\ref{tab:datasets} reports the average number of tokens and the standard deviation for each of the previously described corpora.

\begin{table}[ht!]
    \centering
    \begin{tabular}{ccc}
      \hline
      \textbf{Dataset} & \textbf{Avg. Tokens} & \textbf{Std. Dev.} \\
      \hline
      Raschka 100 & 22.85 & 3.55 \\
      Raschka 1000 & 213.05 & 16.97 \\
      Faculty Chunks & 237.40 & 105.14 \\
      \hline
    \end{tabular}
    \caption{Average number of tokens and corresponding standard deviation for each dataset.}
    \label{tab:datasets}
\end{table}

\subsection{Results}
\label{sec:results}

\begin{table*}[ht!]
  \centering
  \begin{minipage}{0.48\linewidth}
    \centering
    \begin{tabular}{cccc}
      \hline
      \textbf{Model} & \textbf{Param. count} & \textbf{Emb. length} \\
      \hline
      GPT-2        & 137 M & 768 \\
      Llama 3.2 1B & 1.24 B & 2048 \\
      Llama 3.2 3B & 3.21 B & 3072 \\
      Llama 3.1 8B & 8.03 B & 4096 \\
      \hline
    \end{tabular}
    \caption{Overview of the models used in the experiments, including their number of parameters and embedding dimension.}
    \label{tab:models}
  \end{minipage}
  \hfill
  \begin{minipage}{0.48\linewidth}
    \centering
    \begin{tabular}{cccc}
      \hline
      \textbf{Model} & \textbf{Avg. Acc} & \textbf{Reconstructed} \\
      \hline
      GPT-2        & 0.67 & 11 / 20 \\
      Llama 3.2 1B & 0.90 & 15 / 20 \\
      Llama 3.2 3B & 0.87 & 17 / 20 \\
      Llama 3.1 8B & 1.00 & 20 / 20 \\
      \hline
    \end{tabular}
    \caption{Results on the Raschka Blog corpus with chunks of 100 characters.}
    \label{tab:raschka_blog_100}
  \end{minipage}
\end{table*}

\begin{table*}[ht!]
  \centering
  \begin{minipage}{0.48\linewidth}
    \centering
    \begin{tabular}{cccc}
      \hline
      \textbf{Model} & \textbf{Avg. Acc} & \textbf{Reconstructed} \\
      \hline
      GPT-2        & 0.13 & 0 / 20 \\
      Llama 3.2 1B & 0.60 & 1 / 20 \\
      Llama 3.2 3B & 0.85 & 7 / 20 \\
      Llama 3.1 8B & 1.00 & 20 / 20 \\
      \hline
    \end{tabular}
    \caption{Results on the Raschka Blog corpus with chunks of 1000 characters.}
    \label{tab:raschka_blog_1000}
  \end{minipage}
  \hfill
    \begin{minipage}{0.48\linewidth}
    \centering
    \begin{tabular}{cccc}
      \hline
      \textbf{Model} & \textbf{Avg. Acc} & \textbf{Reconstructed} \\
      \hline
      GPT-2        & 0.27 & 0 / 20 \\
      Llama 3.2 1B & 0.68 & 3 / 20 \\
      Llama 3.2 3B & 0.89 & 3 / 20 \\
      Llama 3.1 8B & 1.00 & 20 / 20 \\
      \hline
    \end{tabular}
    \caption{Results on the Faculty of Engineering corpus.}
    \label{tab:fing}
  \end{minipage}
\end{table*}

Tables \ref{tab:raschka_blog_100} and \ref{tab:raschka_blog_1000} report the average accuracy and the proportion of perfectly reconstructed chunks for the Raschka blog corpus, using chunk sizes of 100 and 1000 characters, respectively.
Table \ref{tab:fing} presents the same metrics for the Faculty of Engineering corpus.

\begin{figure}[ht!]
  \centering
  \includegraphics[width=\linewidth]{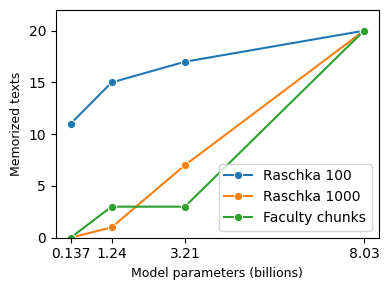}
  \caption{Number of memorized texts as a function of model size across different datasets.}
  \label{fig:graph_size}
\end{figure}

We observe that the largest model, Llama 3.1 8B, successfully reconstructed all sequences across all datasets.
However, model size plays an important role in the ability to reconstruct the original texts.
As shown in Figure~\ref{fig:graph_size}, there is a clear correlation between model size and the proportion of correctly reconstructed texts across all datasets.

\begin{figure}[ht!]
  \centering
  \includegraphics[width=\linewidth]{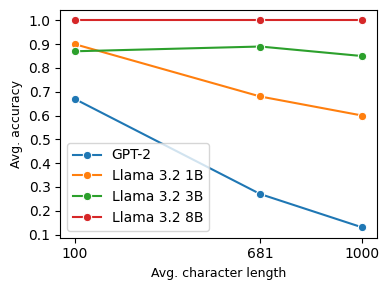}
  \caption{Average accuracy as a function of average character length across different models.}
  \label{fig:graph_length}
\end{figure}

\begin{figure}[ht!]
  \centering
  \includegraphics[width=\linewidth]{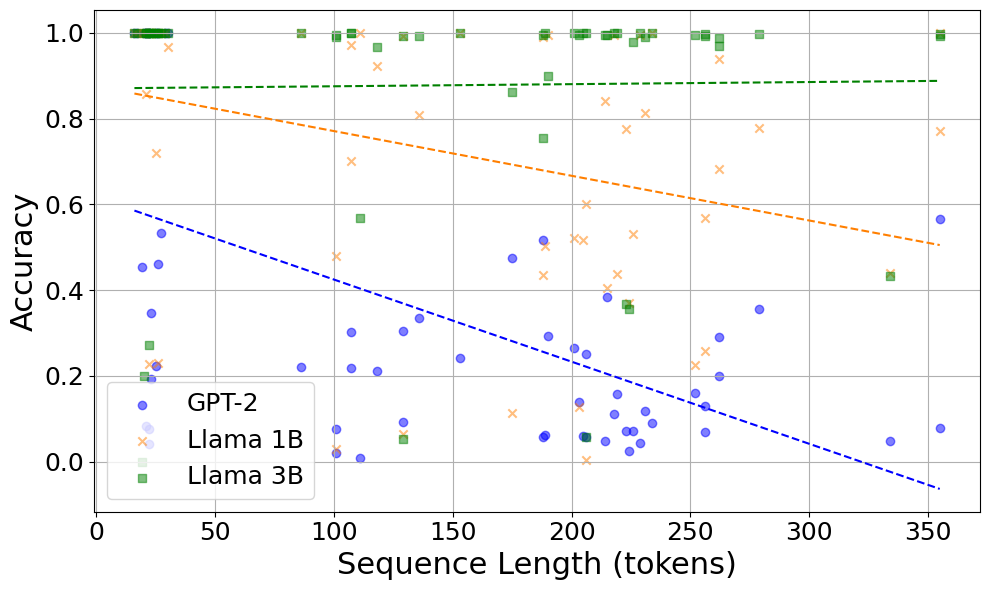}
  \caption{Accuracy as a function of sequence length for GPT-2 and Llama 3.2 1B across all datasets. Each point represents a single evaluation instance. Dashed lines indicate linear trend lines fitted to the data.}
  \label{fig:pointcloud}
\end{figure}

There is also a clear relationship between sequence length and the average accuracy of the smaller models.
Both GPT-2 and Llama 3.2 1B exhibit significant performance degradation on longer sequences, as shown in Figure~\ref{fig:graph_length}.
Following this analysis, Figure~\ref{fig:pointcloud} presents a point cloud combining all datasets for the smaller models. A clear trend emerges: both models show decreasing accuracy as the token count increases, reinforcing the previous observation.

In addition to these experiments, we also tested the reconstruction of Spanish tweets from the HUrtful HUmour (HUHU) shared task~\cite{labadie-tamayo2023everybody}.
These tweets average around 130 characters and present the particular challenge of containing hurtful language or conveying prejudice towards minority groups.
Nonetheless, the model was still able to reconstruct them perfectly, demonstrating robustness even in the presence of offensive content.

When comparing our results to similar work like 500xCompressor~\cite{li2024500xcompressorgeneralizedpromptcompression}, we observe that while we achieve perfect reconstruction of \textasciitilde200-token sequences from a single memory token, under similar conditions (compressing sequences of similar length into a single token using LLaMA 3 8B), they report a Rouge-l-f score of around 0.6, requiring multiple tokens to improve results.
This is not surprising, given that their method relies on an encoder to generate the compressed tokens. However, our results highlight the potential of LLMs to reconstruct text exactly using just one optimized vector.

\section{Conclusions and future work}

We have presented a method for obtaining sentence embeddings from arbitrary text sequences using LLMs and demonstrated that, with models of at least 8B parameters, the original text can be reconstructed using the same LLM, without modifying its weights.

We observed that Llama 3.1 8B was capable of perfectly reconstructing all the sequences.
However, smaller models were less robust and had greater difficulty reconstructing longer sequences, suggesting that, as in many other tasks, model scale plays an important role in performance.

This phenomenon suggests numerous potential applications and directions for future research.

One potential application is in Retrieval-Augmented Generation (RAG)~\cite{gao2024retrievalaugmented}.
Memory tokens could be trained for each chunk.
After the most relevant chunks are obtained in the retrieval step, instead of appending the full text of the retrieved chunks to the prompt, their corresponding embeddings could be used, significantly reducing the number of tokens required in the prompt.

However, further work is needed to find a way to use these embeddings within the LLM for purposes beyond merely reconstructing the original text.
One possible direction is fine-tuning the model with examples where queries are answered using the correct memory tokens, allowing it to learn how to utilize them effectively.

Another important direction for future work is evaluating the effectiveness of the generated sentence embeddings in various downstream tasks, such as classification and retrieval, and comparing their performance with existing methods.

This work also opens the door to exploring LLMs for compression and decompression of information, as memory tokens effectively store entire sequences in compact representations.
Additionally, our experiment with hurtful tweets shows that these embeddings can incite the models to generate harmful content, raising concerns about their potential use in adversarial attacks.
Finally, studying the mechanistic interpretability of LLMs when processing these embeddings could provide deeper insights into why this phenomenon occurs, ultimately contributing to a better understanding of how these models internally represent and retrieve information.

\section{Limitations}

There are some limitations to the work presented in this paper, which we outline below.

This method of generating embeddings is computationally expensive, as it requires backpropagating through the entire network to compute gradients and update the embedding (see Appendix~\ref{appendix} for a training time analysis).
This makes it significantly more demanding than other approaches.
For instance, BERT-based models can generate sentence embeddings with just a forward pass.

This also imposes hardware constraints on running the experiments.
We conducted our experiments using the ClusterUY infrastructure~\cite{clusteruy}, with limited access to an NVIDIA A100 GPU.
As a result, we were unable to run experiments with larger models beyond those presented in Section \ref{sec:results}.

Another limitation of the phenomenon described is that, without fine-tuning or any modifications beyond adding the new embedding, we were unable to use the stored information for tasks other than reconstructing the original sentence.
Intuitively, we believe that LLMs should be capable of effectively use these embeddings for tasks such as Question Answering, without requiring full text reconstruction. However, achieving this may require a fine-tuning step.

This work serves as a demonstration of an interesting phenomenon in LLMs, but further research is essential to explore practical applications.

\section*{Acknowledgments}

This paper has been funded by ANII (Uruguayan Innovation and Research National Agency), Grant No. \texttt{POS\_FMV\_2023\_1\_1012622}.
The experiments presented in this paper were carried out using ClusterUY (site: \url{https://cluster.uy}).

\bibliography{custom}

\appendix

\section{Training Time Analysis}
\label{appendix}

In this appendix, we present an analysis of the time required either to converge (i.e. to generate the expected output exactly) or to reach the maximum number of iterations (3000).
All reported measurements were obtained from experiments run on an NVIDIA A100 GPU, to facilitate reproducibility and comparison.

Table~\ref{tab:time} reports the average number of iterations per second (computed from a sample of 5 runs) and the average total number of iterations to convergence (computed across all samples in the corpus), for the smallest and largest Llama models used, across  the different sequence lengths from the Raschka corpus.

\begin{table}[ht!]
    \centering
    \scalebox{0.8}{
    \begin{tabular}{cccc}
      \hline
      \textbf{Model} & \textbf{Seq. length} & \textbf{Avg. Its/s}& \textbf{Avg. Total Its.} \\
      \hline
      Llama 3.2 1B & 100 & 16.10 & 1120.35 \\
      Llama 3.1 8B & 100 & 8.14 & 146.52 \\
      \hline
      Llama 3.2 1B & 1000 & 6.14 & 2987.90 \\
      Llama 3.1 8B & 1000 & 6.04 & 707.55 \\
      \hline
    \end{tabular}
    }
    \caption{Average iterations per second (Its/s) over 5 runs and average total iterations (Total Its.) until convergence across different sequence lengths and models.}
    \label{tab:time}
\end{table}

It can be observed that the time per iteration varies depending on the input sequence length.
For instance, for Llama 3.1 8B, we observe 8.14 iterations per second on average for sequences of 100 characters, and 6.04 iterations per second for sequences of 1000 characters.

As expected, the number of iterations required per example also depends on the sequence length.
In the 100-character case, it takes an average of 146.52 iterations to perfectly reconstruct the original text (around 18 seconds in total), whereas for 1000-character sequences, the average rises to 707.55 iterations (around 117 seconds in total). 
Although this is significantly higher, it remains well below the maximum limit of 3000 iterations we imposed.

While the smaller model achieves a higher iteration rate, it typically requires more iterations to converge, especially with longer sequences, where it often fails to do so within the 3000-iteration limit.
Surprisingly, with the longer sequences, the iteration rate is nearly identical between the two models, suggesting that sequence length has a noticeable impact on iteration speed. 

\begin{figure}
  \centering \includegraphics[width=\linewidth]{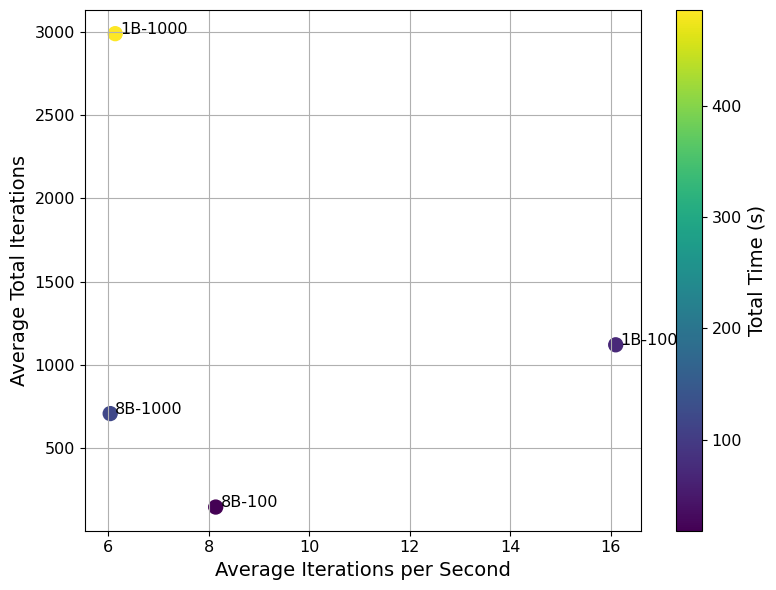}
  \caption{Trade-off between iteration speed and total iterations for different Llama models and input lengths. Color indicates total time to convergence (in seconds).}
  \label{fig:time-scatter}
\end{figure}

Figure~\ref{fig:time-scatter} illustrates the trade-off between iteration speed and the number of iterations required for convergence.
The larger model is faster overall for both sequence lengths, thanks to its ability to converge in fewer steps.

\end{document}